\documentclass[10pt,twocolumn,letterpaper]{article} 

\usepackage{avss}
\usepackage{times}
\usepackage{epsfig}
\usepackage{graphicx}
\usepackage{amsmath}
\usepackage{amssymb}
\usepackage{array}

   \avssfinalcopy 

\def\avssPaperID{3} 

 \ifavssfinal\fi
\begin{document}

\title{ResnetCrowd: A Residual Deep Learning Architecture for Crowd Counting, Violent Behaviour Detection and Crowd Density Level Classification}

\author{Mark Marsden, Kevin McGuinness, Suzanne Little, Noel E. O'Connor\\
Insight Centre for Data Analytics \\
Dublin City University, Ireland\\
{\tt\small mark.marsden@insight-centre.org \{kevin.mcguinness,suzanne.little,noel.oconnor\}@dcu.ie}
}

\maketitle

\begin{abstract}
In this paper we propose ResnetCrowd, a deep residual architecture for simultaneous crowd counting, violent behaviour detection and crowd density level classification. To train and evaluate the proposed multi-objective technique, a new 100 image dataset referred to as Multi Task Crowd is constructed. This new dataset is the first computer vision dataset fully annotated for crowd counting, violent behaviour detection and density level classification. Our experiments show that a multi-task approach boosts individual task performance for all tasks and most notably for violent behaviour detection which receives a 9\% boost in ROC curve AUC (Area under the curve).  The trained ResnetCrowd model is also evaluated on several additional benchmarks highlighting the superior generalisation of crowd analysis models trained for multiple objectives.

\end{abstract}

\section{Introduction}
The automated analysis of highly congested, highly varied crowded scenes is a challenging vision task that has received a lot of attention in recent years from both the computer vision research community and private industry alike. With the rapid increase in global population seen over the last century, particularly in urban
areas, highly congested crowds have become a part of daily life that present enormous
challenges to maintaining public safety. Every year dozens of people die in urban areas
due to stampedes and crushes, such as the New Year's Eve 2014 stampede in Shanghai,
where 36 people tragically died. This highly preventable loss of life could potentially be avoided with better
analysis and understanding of crowd behaviour and congestion levels in our cities.

Work to date in the crowd analysis area has focused on developing task specific systems which perform a single analysis task such as crowd counting \cite{zhang2016single}, crowd behaviour recognition \cite{Kang2015}, crowd density level classification \cite{Fu2015} and crowd behaviour anomaly detection \cite{Li2014a}.  It has been shown in other domains such as facial analysis \cite{Ranjan2016}  that learning correlated tasks simultaneously can boost individual task performance. However, a multi-objective learning approach to crowd analysis has yet to be fully investigated due largely to the lack of an appropriately labelled multi-task dataset.

In this paper we propose a residual deep learning framework for simultaneous crowd counting, violent behaviour recognition and crowd density level classification. We refer to this architecture as ResnetCrowd. Residual deep learning architectures have been shown to achieve state-of-the-art performance in both image recognition and object detection tasks \cite{he2016deep}.  We propose a new Multi Task Crowd dataset to train this network. This new dataset is the first computer vision dataset fully annotated for crowd counting, violent behaviour detection and density level classification. The core contributions of this paper include:

\begin{enumerate}
  \item  The construction of a 100 image dataset fully labelled for crowd counting, violent behaviour detection and crowd density estimation,
  \item  A deep, residual neural network architecture for simultaneous crowd counting, violent behaviour detection and crowd density estimation,
  \item  A quantitative demonstration of the benefits of multi-objective crowd analysis systems.

\end{enumerate}
The remainder of the paper is organised as follows: Section 2 presents a review of the related work. Section 3 describes the construction of the Multi Task Crowd dataset. Section 4 details the proposed ResnetCrowd neural network architecture while section 5 presents a comprehensive set of experiments which highlight the benefits of multi-objective crowd analysis.
\section{Related Work}

Multi-objective approaches to crowd analysis have shown some initial promise, such as the work of Hu \etal \cite{hu2016dense}, who showed that the inclusion of density level classification increased the robustness of their crowd counting system. To date, no crowd analysis technique has been developed which encompasses both behaviour recognition and crowd counting/scene occupancy. The benefits of multi task learning have been successfully demonstrated in areas such as facial analysis \cite{Ranjan2016}, head pose estimation \cite{yan2016multi} and speech recognition \cite{seltzer2013multi}. The following discussion reviews existing work in each crowd analysis task domain. 

\textbf{Crowd Counting} Crowd counting algorithms attempt to produce an accurate estimation of the true number of people present in an image of a crowded scene. The emergence of deep neural network techniques such as convolutional neural networks and the availability of high density, high variation crowd counting datasets such as UCF\_CC\_50 \cite{idrees2013multi} has resulted in state-of-the-art crowd counting techniques such as the work of Marsden \etal \cite{mark_count}. The majority of recent approaches train a crowd counting regressor to directly map pixel values to a single count estimate \cite{Zhang2015,Hu2016}, however pixel-wise heatmap based counting  has been shown to improve crowd counting performance for challenging, highly congested scenes \cite{zhang2016single}.

\textbf{Crowd Behaviour Recognition} 
Crowd behaviour recognition techniques attempt to categorise the behaviour observed in an image or video of a crowded scene. Crowd behaviour classification should be seen as a distinct task from human action recognition which typically focuses on a single subject. Hand crafted  inter-frame motion features were used by Hassner \etal to detect violent crowd behaviour \cite{Hassner2012}. This type of approach relies upon a contiguous sequence of frames and cannot classify still images. 
General purpose crowd behaviour concept detection has been investigated  by Kang \etal \cite{Kang2015} whose technique produces probability scores for a range of crowd behaviour concepts ranging from the very innocuous (``walking, skating'') to highly salient concepts (``Fight'', ``Mob''). Detecting concepts such as "walking" and "skating" is useful for video retrieval and image captioning systems but is of little use to the security community. This approach again relies on inter-frame motion features.

\textbf{Crowd Density Level Estimation} 
Crowd density level refers to the level of crowd congestion observed in a crowded scene. This aspect of a crowded scene is typically expressed either as a discrete (0-N) or continuous value (0.0-1.0). Texture analysis features were used by Wu \etal  to produce a continuous density level estimate \cite{wu2006crowd}. More recently a deep convolutional neural network was used by Fu \etal \cite{Fu2015} for discrete density level classification. The main issue with this task is the level of ambiguity associated with a given density level estimate.  There is no set scheme across datasets for assigning density level labels and the specific associated meanings. The most transparent scheme possible is one where discrete density level labels are inferred directly from true crowd count values, producing a histogram style distribution with subjectivity and human error minimised.

\section{Multi Task Crowd Dataset}
The core objective of the Multi Task Crowd Dataset is to produce a set of images suitable for training and validating a model for simultaneous crowd counting, violent behaviour recognition and crowd density level classification. The dataset and associated experiments  evaluate single frame crowd analysis performance, a real-world scenario that must be considered. Violent behaviour recognition is targeted because of its importance to security personnel. Discrete crowd density level classification is chosen because of the lack of subjectivity involved and because discrete density level labels can be automatically inferred from crowd count ground truths. With all this in mind the following criteria were followed when constructing the dataset.

\begin{enumerate}
  \item Significant variation in scene content
  \item  An even split between images of violent and non-violent behaviour
  \item Significant variation in crowd size
\end{enumerate}

A publicly available dataset which meets these requirements has not been produced to date due to the expensive and time consuming nature of image annotation. Tasks specific collections such as WWW Crowd \cite{Kang2015} and UCF\_CC\_50 \cite{idrees2013multi} have emerged in recent years and facilitated significant progress in the behaviour recognition and crowd counting areas respectively.

The most efficient way to produce the desired Multi Task Crowd dataset is to apply new labels for additional analysis tasks to an existing dataset. The UCF\_CC\_50 dataset contains high quality images of large crowds, but with little variation in terms of behaviour and scene content. On the other hand, the WWW Crowd dataset contains 10,000 video clips of crowds annotated for 94 crowd behaviour concepts. This dataset contains significant variation in behaviour and crowd size and is therefore used to construct Multi Task Crowd.

\begin{figure}[h!]
	\centering

    \includegraphics[width=0.2\textwidth]{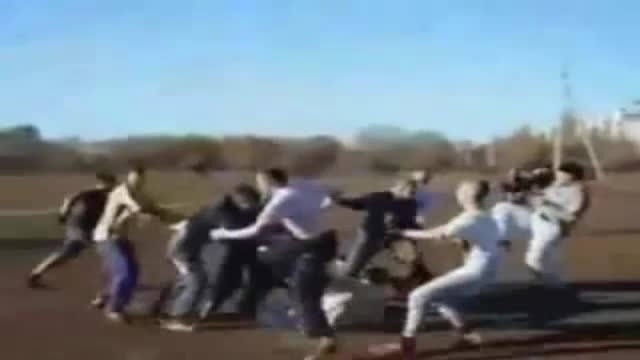}
    \includegraphics[width=0.2\textwidth]{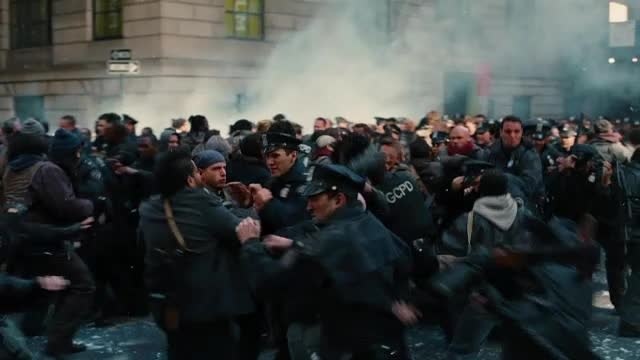}

    \caption{Violent behaviour images used in the Multi Task Crowd dataset}
    	\label{violent_samples}
\end{figure}

\begin{figure}[h!]
	\centering
	\includegraphics[width=0.2\textwidth]{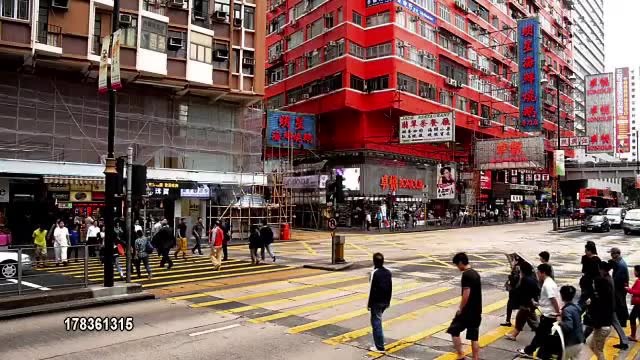}
      \includegraphics[width=0.2\textwidth]{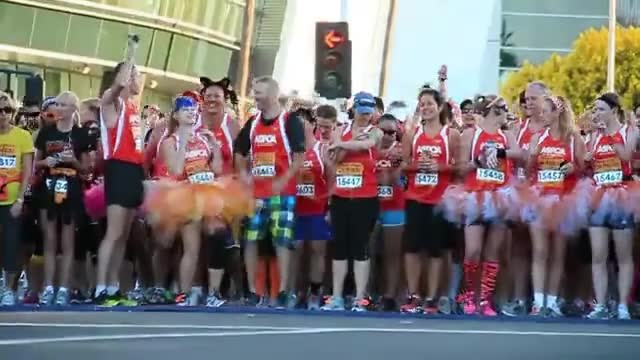}

	\caption{Non-violent behaviour images used in the Multi Task Crowd dataset}
    \label{non_violent_samples}
\end{figure}

A set of violent behaviour images is gathered by finding all WWW Crowd training clips where either the ``Fight'' or ``Mob'' concepts are present. The first frame of each clip is extracted and a 50 image subset is produced. High variation in crowd size is achieved during the selection process by observing each frame and ensuring there is a significant quantity of low (0-50 people), medium (50-150 people)  and high (150+ people) crowd density images. A similar process is then carried out for WWW crowd clips where the ``Fight'' and ``Mob'' concept are not present, with another 50 image subset extracted. Sample images from the violent and non-violent subsets are shown in figures \ref{violent_samples}  and \ref{non_violent_samples}.

\begin{figure}[h!]
	
    \centering
    \includegraphics[width=0.30\textwidth]{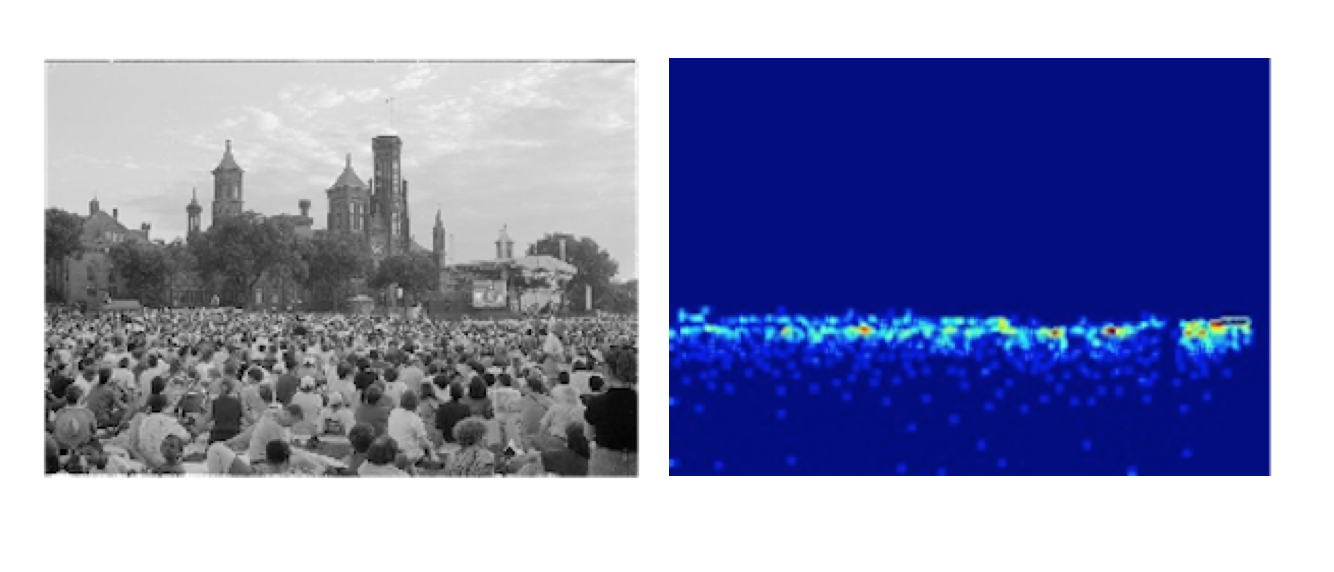}

	\label{heatmap_example}
    \caption{Sample crowd heatmap produced for a given crowd image using the method of Zhang \etal \cite{zhang2016single}. The Jet colourmap has been applied for visualisation purposes }.
\end{figure}

\begin{figure}[h!]
	\centering
	\includegraphics[width=0.30\textwidth]{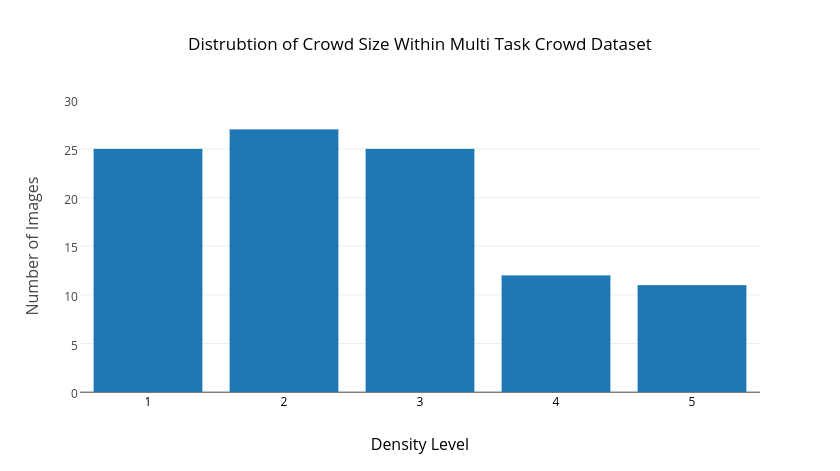}
	\caption{Crowd size distribution within the Multi Task Crowd dataset}
    \label{density_distribution}
\end{figure}

Combining these two sets produces a 100 frame dataset evenly split between violent and non-violent behaviour with high variation in crowd size and scene content. Taking the ``Mob'' and ``Fight'' behaviour labels from the WWW crowd annotation data provides the violent behaviour detection ground truth for our dataset. Additional labels are then applied to these images for crowd counting and density level classification. Crowd counting labels are applied by marking the head of each person in a given image with a single pixel, in a manner similar to the UCF\_CC\_50 dataset, with the total number of marks equal to the true person count. A crowd heatmap is also produced for each image using the approach of Zhang \etal \cite{zhang2016single}, which takes head annotation data and produces a smooth crowd heatmap where the integral is equal to the crowd count. These crowd heatmap images are used to train a pixel-wise approach to crowd counting which will be compared to regression-based counting. All ground truth crowd heatmaps are downscaled to $160 \times 90$ in order to match the predicted heatmap resolution of the ResnetCrowd model. Figure \ref{heatmap_example} illustrates the crowd heatmap produced for a given crowd image using the method of Zhang \etal \cite{zhang2016single}. Discrete density level labels are then inferred from these overall person counts for each image using the scheme proposed in table \ref{density_scheme}. The distribution of crowd sizes within the produced dataset is highlighted in figure \ref{density_distribution}. 

\begin{table}[h!]
\centering{%
\begin{tabular}{m{2.9cm} m{2cm} m{2cm} }
\hline
Density Level Label & Minimum Count & Maximum Count \\ \hline 
1                   & 0             & 20            \\ \hline
2                   & 21            & 50            \\ \hline
3                   & 51            & 100           \\ \hline
4                   & 101           & 200           \\ \hline
5                   & 201          & 201+          \\ \hline
\end{tabular}
}

\caption{Density level annotation scheme used during the construction of the Multi Task Crowd dataset}
\label{density_scheme}
\end{table}

The final dataset thus consists of 100 images, along with the following annotation data for each: a discrete density level in the range 1-5, an overall crowd count value, head locations for each person in the scene as well as binary labels indicating the presence or absence of the ``Mob'' and ``Fight'' behaviour concepts. Benchmarking for all tasks is carried out on this dataset using a 5 fold cross-validation, with care taken to ensure each fold is representative of the overall set.

\section{ResnetCrowd}
The proposed ResnetCrowd architecture is based upon the Resnet18 network of He \etal \cite{he2016deep}. The initial 5 convolutional layers of Resnet18 as well as the interleaved batch normalisation \cite{ioffe2015batch} layers and skip connections form the primary module of our network which is illustrated in figure 5. The max pooling layer which follows the first convolutional layer of Resnet18 is removed in order to keep suitably large feature maps for pixel-wise crowd counting.  This initial portion of the network is initalised with weights from a Resnet18 network trained on the ImageNet dataset. Relu (Rectified Linear Unit) activations are applied after each convolutional layer. Resnet18 was chosen for its low parameter count, high performance on image classification tasks and fast convergence \cite{he2016deep}. The feature map average pooling step used in Resnet-like architectures allows the fully connected layers used for classification to contain significantly fewer parameters. This overall reduction in parameters enables small dataset problems such as multi-task crowd analysis to be successfully trained. Only the initial 5 convolutional layers were included due to memory limitations on the hardware used.

\begin{figure}[h!]
	\centering
	\includegraphics[width=0.15\textwidth]{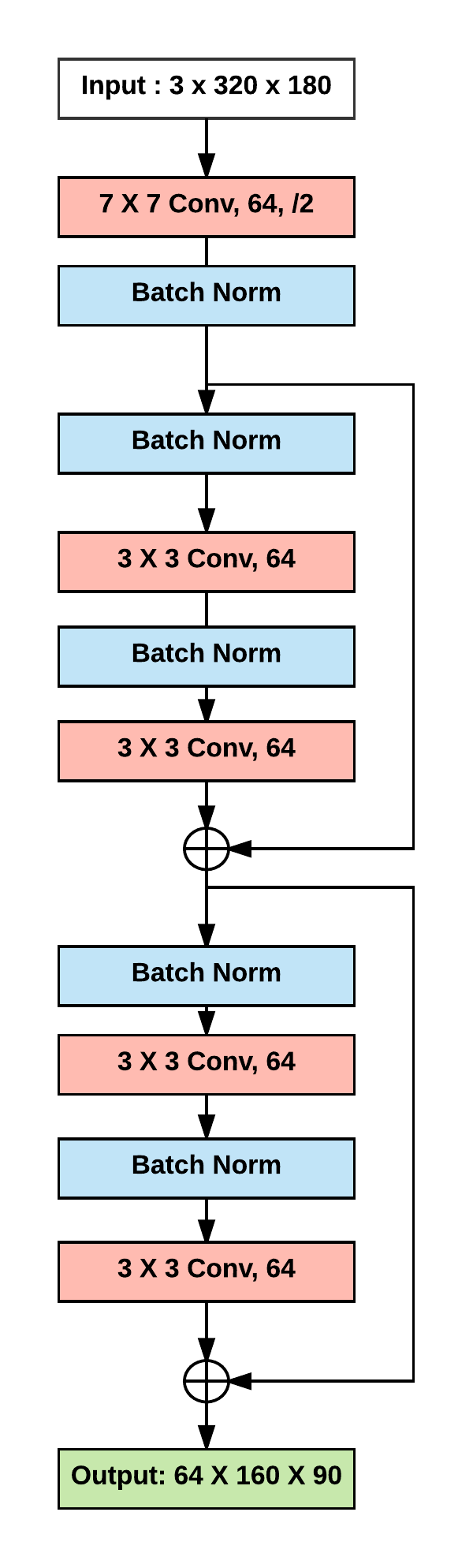}
	\label{bla}
    \caption{Primary module of ResnetCrowd}
\end{figure}

Following these 5 convolutional layers a set of task specific layers are added to ResnetCrowd. First the pixel-wise crowd counting is performed using the  \textit{CountingHeatmap} convolutional layer which performs a 1 $\times$ 1 convolution to output an estimated crowd density heatmap. Second the 64 feature maps produced by our initial network are average pooled to produce a shared, 64 dimensional representation from which classification tasks can be trained. Task specific fully connected layers for regression based crowd counting, violent behaviour detection and density level classification are then added. The weights for these task specific additional layers are initialised using Xavier initalisation \cite{glorot2010understanding}.  The task specific module of our network is illustrated in figure 6. ResnetCrowd is then trained end-to-end by combining these two modules. The total parameter count for the proposed architecture is just 180,934.

\begin{figure}[h!]
	\centering
	\includegraphics[width=0.4\textwidth]{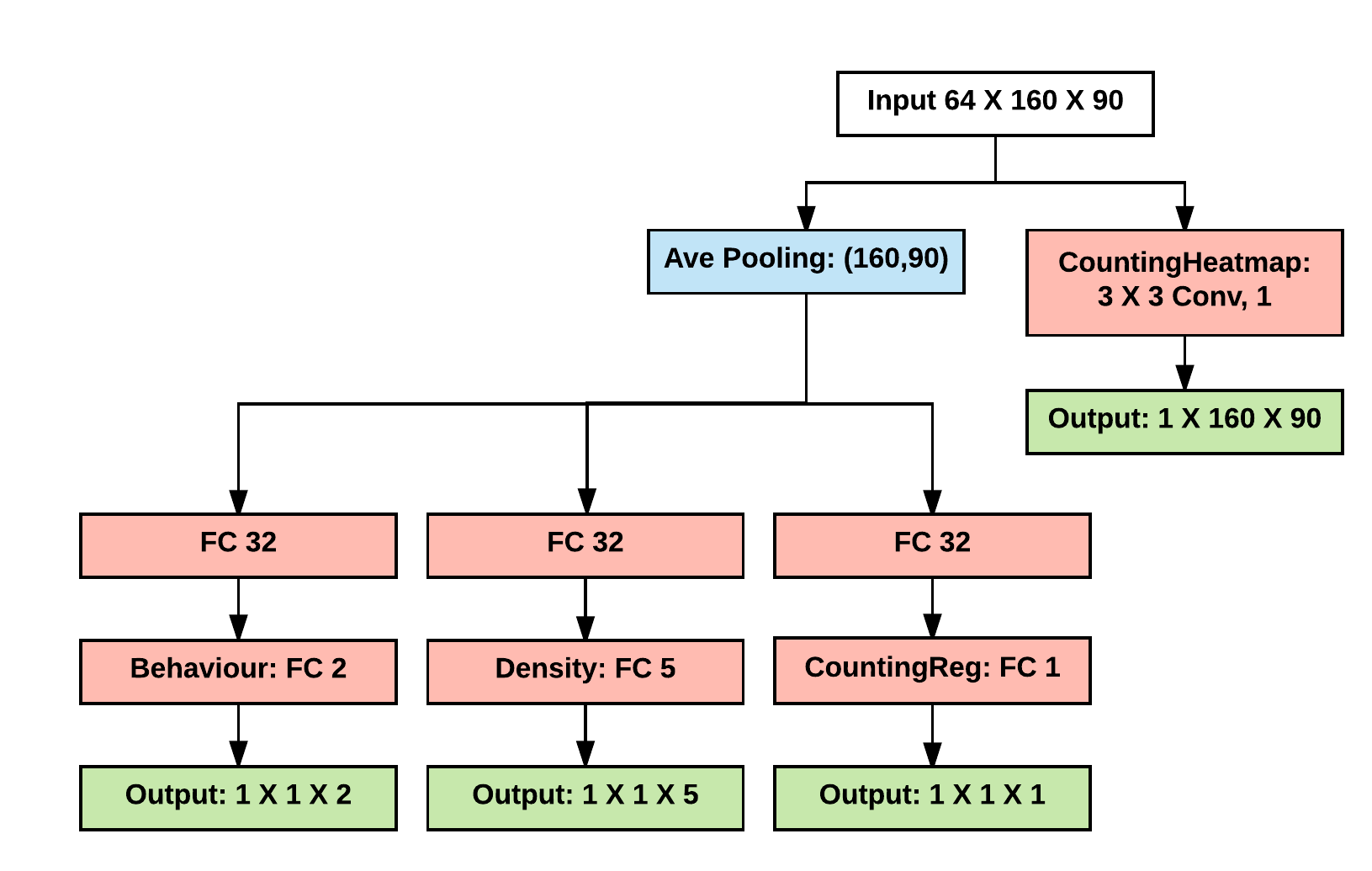}   
	\label{task_arch}
    \caption{Tasks specific module of the ResnetCrowd architecture}
\end{figure}

The ResnetCrowd architecture is trained on the Multi Task Crowd dataset by minimising a loss function which combines losses for each of the 4 supervised outputs. The resolution of all images is halved to $320 \times 180$ to ensure a suitably high batch size is maintained during network optimisation. The AdaGrad optimiser \cite{duchi2011adaptive} was utilised to avoid learning rate selection issues. L2 weight regularisation (i.e. weight decay) was also enforced during training with $\lambda$ set to $1 \times 10^{-4}$. Task specific output activations and loss functions are detailed as follows. 

\textbf{Behaviour Recognition} A sigmoid activation is applied to the output of the \textit{Behaviour} fully connected layer as the objective of this task is to predict probability scores for each behaviour concept (``Mob'' and ``Fight'') individually. Binary cross entropy, given in equation \ref{BCE_behave}, is thus the most appropriate loss to minimise for this task. $\hat{S}_j$ refers to the predicted probability score for concept $j$ while $S_j$ refers to the ground truth scores.
\begin{equation}
L_{\text{Behave}}=-\frac{1}{N}\sum_{j=1}^{N}S_{j}\log(\hat{S_{j}})+(1-S_{j})\log(1-\hat{S_{j}})
\label{BCE_behave}
\end{equation}

\textbf{Density Level Classification}
As the 5 density level labels discussed in section 2 are mutually exclusive a more conventional classification approach is taken for density level classification with a softmax activation applied to the \textit{Density} output and a categorical crossentopy loss (given in equation \ref{den_CCE}) is minimised. $\hat{S}_{ij}$ refers to the predicted probability of category $j$ on example $i$ while $S_{ij}$ refers to the same for the ground truth. 
\begin{equation}
L_{\text{Density}}=-\frac{1}{N}\sum_{i=1}^{N} \sum_{j=1}^5 S_{ij}\log(\hat{S_{ij}})
\label{den_CCE}
\end{equation}

\textbf{Regression Based Crowd Counting}
A Relu activation is applied to the \textit{CountingReg} output to ensure no negative counting estimates are produced. Mean squared error (given in equation \ref{MSE_reg}) is minimised for this task.
\begin{equation}
L_{\text{CountReg}}={\frac{1}{N}\sum_{j=1}^{N}( S_{j}-\hat{S}_{i}  )^{2} }
\label{MSE_reg}
\end{equation}

\textbf{Heatmap Based Crowd Counting}
The heatmap based crowd counting task is trained by comparing a predicted crowd heatmap with the corresponding ground truth heatmap. This can be modeled as predicting the probability of each pixel containing a person. We implement this by applying a sigmoid activation to each pixel of the \textit{CountingHeatmap} output and minimizing the binary cross entropy loss between the predicted and ground truth heatmap. This loss function is given in equation \ref{BCE_count}. At inference time an overall crowd count estimate is produced by integrating an estimated crowd heatmap as is performed in the work of Zhang \etal \cite{zhang2016single}.
\begin{equation}
L_{\text{CountingHeatmap}}=-\frac{1}{N}\sum_{j=1}^{N}S_{j}\log(\hat{S_{j}})+(1-S_{j})\log(1-\hat{S_{j}})
\label{BCE_count}
\end{equation}

The total loss is computed as a sum of 4 individual losses as shown in equation \ref{sum_loss}.

\begin{equation}
L_{\text{Total}}=L_{\text{Behave}}+L_{\text{Density}}+L_{\text{CountReg}}+L_{\text{CountingHeatmap}}
\label{sum_loss}
\end{equation}

\section{Experimental Results}
 We evaluate the proposed ResnetCrowd architecture on the Multi Task Crowd dataset. A 5-fold cross validation is performed for each experiment. The training set used for each dataset fold is augmented with horizontal rotations, doubling the size. All network optimisation and testing is performed using an Nvidia GeForce GTX 970 GPU with batch size set to 40. Our technique is implemented using the Keras neural network API \cite{chollet2015keras} with a tensorflow backend \cite{abadi2015tensorflow}. 

\subsection{Multi vs Single Task Learning}
This section compares the proposed multi-task ResnetCrowd network to single task baseline runs for violent behaviour detection, crowd density level estimation, regression based crowd counting and heatmap based crowd counting. For each single task run the architecture remains identical with only the task specific module altered to contain just the layers used for the given task (e.g. only the \textit{CountingHeatmap} layer remains for the heatmap based counting baseline). Training is performed for 500 epochs per cross validation fold for all runs. The training set order is randomly shuffled between epochs. Mean performance for all runs is shown in table \ref{overall_performance}.  Violent behaviour detection AUC is improved by 9\% to 0.78 while small performance improvements are observed across all other tasks. 

To better examine the effects of  multi-task learning on crowd counting performance the mean absolute error metric is  reported separately for low (0-50 people), medium (50-150 people) and high (150+ people) congestion images in table \ref{counting_breakdown}. For single task runs, regression based counting outperforms heatmap based counting for low density scenes, while heatmap based counting achieves significantly better performance on high density scenes. This overall performance breakdown is altered when we observe the ResnetCrowd run. For both regression and heatmap based counting the performance on low density scenes is boosted at the expense of performance on high density scenes.

\begin{table*}[h!]
\centering
\begin{tabular}{m{5.5cm} m{1.5cm} m{1.7cm} m{1.7cm} m{1.5cm} m{1.5cm} m{1.5cm}}
\hline
Run                                  & Behaviour: mAUC $\uparrow$  & Density: Accuracy $\uparrow$  & Regression Counting: MAE $\downarrow$ & Heatmap Counting: MAE $\downarrow$ \\ \hline
Single Task Behaviour                & 0.71            & N/A                                         & N/A                       & N/A                    \\ \hline
Single Task Density Level Estimation & N/A             & 0.4                                         & N/A                       & N/A                    \\ \hline
Single Task Regression Counting      & N/A             & N/A                                          & 58.4                      & N/A                    \\ \hline
Single Task Heatmap Counting         & N/A             & N/A                                          & N/A                       & 58.6                   \\ \hline
ResnetCrowd                          & \textbf{0.78}   & \textbf{0.42}                              & \textbf{58.3}                      & \textbf{58.4}                   \\ \hline
\end{tabular}
\caption{Performance comparison of the ResnetCrowd architecture with single task baselines}
\label{overall_performance}
\end{table*}

\begin{table*}[h!]
\centering
\begin{tabular}{m{6cm} m{2.5cm} m{3.0cm} m{2.5cm}}
\hline 
Run                               & Low Congestion MAE  $\downarrow$     & Medium Congestion MAE $\downarrow$  & High Congestion MAE $\downarrow$     \\ \hline
Single Task Regression Counting   & 25            & 37          & 217          \\ \hline
Single Task Heatmap Counting      & 49            & \textbf{19} & \textbf{175} \\ \hline
ResnetCrowd: Regression Counting & \textbf{11}            & 39          & 253         \\ \hline
ResnetCrowd: Heatmap Counting     & 21            & 52          & 221         \\ \hline
\end{tabular}
\caption{Crowd counting mean absolute error (MAE) performance for ResnetCrowd}
\label{counting_breakdown}
\end{table*}
   
\subsection{Transfer Learning}
The transfer learning capability of ResnetCrowd is investigated by comparing performance with the state-of-the-art on several task specific benchmarks. The goal of these experiments is to observe how multi-task learning can enhance the generalisation of a model trained on a given dataset (Multi Task Crowd).

\subsubsection{Violent Behaviour Recognition}
A trained ResnetCrowd model is used to perform violent behaviour recognition on the WWW crowd test set \cite{Kang2015}. This set contains 1834 video clips with the goal being to detect the occurrence of 94 crowd behaviour concepts. For this experiment ROC curve performance will be evaluated for just the "Fight" and "Mob" concepts and compared with the state-of-the-art. Concept probability scores are predicted for every 10th frame of a given clip and the mean taken for that clip. The performance of ResnetCrowd is highlighted in table \ref{violent_WWW} and compared with the violent behaviour detection single task baseline as well as the state-of-the-art approach. 

\begin{table}[h!]
\centering
\begin{tabular}{m {1.95cm} m{2cm} m{2cm}}
\hline
Run & Fight AUC $\uparrow$ & Mob AUC $\uparrow$ \\ \hline
Single Task Behaviour & 0.62 & 0.68 \\ \hline
ResnetCrowd & 0.71 & 0.77 \\ \hline
Kang \etal \cite{Kang2015} & \textbf{0.93} & \textbf{0.91} \\ \hline
\end{tabular}

\caption{Crowd behaviour concept detection performance on the WWW crowd test set}
\label{violent_WWW}
\end{table}

ResnetCrowd significantly outperforms the single task baseline run despite being trained on an identical set of images. When compared to the state-of-the-art technique, AUC performance only falls by 15\% for "Mob" and 20\% for "Fight", which is impressive considering only 80 frames were used for training compared to the several million frames available to Kang \etal to train their original 94 concept model.

\subsubsection{Crowd Counting}
The ResnetCrowd model is used to perform crowd counting on the UCF\_CC\_50 dataset. This highly challenging 50 image dataset contains crowds which vary in size from 45 to 4500 people. Table \ref{UCF_count} compares counting performance of ResnetCrowd (both the regression and heatmap counting outputs) with single task baselines as well as the leading techniques.

\begin{table}[h!]
\centering
\begin{tabular}{lll}
\hline
Run                & MAE $\downarrow$  & MSE $\downarrow$  \\ \hline
Single Task Regression   & 1128 & 1478 \\ \hline
Single Task Heatmap      & 989  & 1346 \\ \hline
ResnetCrowd : Regression & 1150 & 1497 \\ \hline
ResnetCrowd: Heatmap     & 896  & 1267 \\ \hline
Zhang \etal  \cite{zhang2016single}            & 377  & 509  \\ \hline
Marsden \etal \cite{mark_count}        & \textbf{338}  & \textbf{425} \\ \hline
\end{tabular}
\caption{Crowd counting performance on the UCF\_CC\_50 dataset}
\label{UCF_count}
\end{table}

ResnetCrowd improves heatmap based counting performance by 9\% compared to the respective single task baseline. Regression based counting performs poorly on all runs, highlighting the advantages of heatmap based counting. The use of fully connected layers in regression based counting networks requires the input image to be resized to a fixed resolution ($ 320 \times 180$ in this case), while heatmap based counting allows the original image resolution to be maintained. This forced resampling is one of the major limitations of regression based counting. The inferior performance of ResnetCrowd when compared to the leading techniques can largely be attributed to the network being trained on lower density crowd images taken from the Multi Task Crowd dataset. It is also important to note that no fine tuning was performed for the UCF\_CC\_50 dataset. However generalisation and crowd counting performance is clearly improved through the use of a mulit-task learning approach.

\subsubsection{Crowd Behaviour Anomaly Detection}
A trained ResnetCrowd model is used to perform crowd behaviour anomaly detection on the UMM dataset \footnote{http://mha.cs.umn.edu/Movies/Crowd-Activity-All.avi}. Removing all task specific layers from our architecture other than the average pooling layer leaves a network with a 64 dimensional vector output. We investigate how successfully these single-frame features can be used for crowd behaviour anomaly detection by passing each frame of the UMN dataset through ResnetCrowd and training a Gaussian mixture model to perform outlier detection using the generated vectors. We also compare how features produced using our ResnetCrowd architecture perform versus those from our single task violent behaviour recognition baseline. Results from this experiment are shown in table \ref{anomaly_detection}. ResnetCrowd significantly outperforms the single task behaviour recognition baseline. This result highlights the potential of the feature representations produced through multi-task crowd analysis.   While these results are far from state-of-the-art for this task it is important to note that ResnetCrowd does not utilise any inter-frame motion features like the leading techniques \cite{mark_count,Li2014a}. These leading approaches also apply hand-crafted features specifically engineered for this task unlike the multi-purpose features learned by ResnetCrowd.

\begin{table}[h!]
\centering
\begin{tabular}{m{4cm} m{1cm}}
\hline 
Run                               & AUC   $\uparrow$         \\ \hline
Single Task Behaviour    & 0.73                  \\ \hline
ResnetCrowd     & 0.84           \\ \hline
Marsden \etal \cite{Marsden2016} & 0.92                   \\ \hline
Li \etal  \cite{Li2014a}    & \textbf{0.99}         \\  \hline
\end{tabular}
\caption{Crowd behaviour anomaly detection performance on the UMN dataset}
\label{anomaly_detection}
\end{table}


\section{Conclusions}
In this paper we have demonstrated the benefits of multi task crowd analysis through the development of a residual learning approach to simultaneous crowd counting, violent behaviour detection and crowd density level classification. A 100 image dataset has been constructed to evaluate the performance of the proposed multi task architecture. Future work will look to include unsupervised learning techniques to overcome the lack of labelled crowd data and further increase model generalisation.

\section{Acknowledgments}
This paper is based on research supported by Science Foundation Ireland under grant number SFI/12/RC/2289

{\small
\bibliographystyle{unsrt}
\bibliography{egbib}
}

\end{document}